\pdfoutput=1

\documentclass[11pt]{article}

\usepackage[]{acl}

\usepackage{times}
\usepackage{latexsym}

\usepackage[T1]{fontenc}

\usepackage[utf8]{inputenc}

\usepackage{microtype}
\usepackage{graphicx}
\usepackage{multirow}
\usepackage{algorithm}
\usepackage{algpseudocode}
\usepackage{amsmath} 
\usepackage{multirow}
\usepackage{booktabs} 
\usepackage{enumitem} 
\usepackage{subcaption} 
\usepackage{wrapfig}

\title{
Joint Training for Selective Prediction
}

\author{{Z}haohui {L}i  \and {R}ebecca {J}. {P}assonneau \\ {D}epartment of {C}omputer {S}cience and {E}ngineering\\
        {P}ennsylvania {S}tate {U}niversity \\  \{zjl5282,  rjp49\}@psu.edu}

\begin{document}
\maketitle
\begin{abstract}

Classifier models are prevalent in natural language processing (NLP), often with high accuracy. Yet in real world settings, human-in-the-loop systems can foster trust in model outputs and even higher performance. 
Selective Prediction (SP) methods determine when to adopt a classifier's output versus defer to a human. 
Previous SP approaches have addressed how to improve softmax as a measure of model confidence, or have developed separate confidence estimators. 
One previous method involves learning a deferral model based on engineered features.
We introduce a novel joint-training approach that simultaneously optimizes learned representations used by the classifier module and a learned deferral policy.
Our results on four classification tasks demonstrate that joint training not only leads to better SP outcomes over two strong baselines, but also improves the performance of both modules. 

\end{abstract}


\section{Introduction}

One of the key questions in application of natural language processing (NLP) systems to real-world problems is how best to provide automated support for decisions that humans find laborious, have difficulty making in a consistent and therefore fair manner, and that have high social impact. In areas such as education, human expertise can be best utilized when experts are relieved of more routine decisions that can be reliabily automated. Yet, concerns have been raised about how machines and humans can become better collaborators in education \cite{cardona2023artificial}.  Educators can learn more about students' misconceptions when they answer questions in their own words, so automated support could shift human effort from grading towards design of effective curricula. Among the diverse types of human-in-the-loop systems~\cite{mosqueira-rey_human-loop_2022}, selective prediction on classification tasks investigates when and how to defer a decision to a human expert~\cite{Bondi2022, Geifman2017,xin2021art}. Previous work on selective prediction has focused on differentiating between high versus low model confidence, to develop a deferral policy after the fact. In contrast, we develop an approach to jointly train a classifier and a deferral policy, which we test on diverse datasets of open-ended STEM questions.

Previous work on selective prediction (SP) has investigated ways to improve the decision to defer, based on better information about the likelihood that a classifier's prediction is correct. One type of information comes from close examination of the softmax output that provides a probability distribution over the output classes~\cite{hendrycks_baseline_iclr2017,xin2021art}. Machine learned methods have been proposed to learn a model that independently estimates the classifier confidence, referred to as calibration~\cite{garg_will_2021,kamath_selective_2020,Varshney2022settings}. Another work learned a logistic regression deferral classifier using softmax output combined with features representing difficulty for the model of a set of examples (e.g., a particular STEM assessment question)~\cite{li2023learning}.  To avoid confusion with the main classifier in the SP setting (CL), we will refer to a deferral classifier as a deferral policy (DP). Often, a deferral policy is trained on a dataset distinct from the one used to train the CL. The key issue is that the supervision signal for the deferral policy depends on the CL producing its decisions. As a result, all previous methods have relied on a previously trained CL.  Here we jointly train the classifier and the deferral policy, which we refer to as Joint Training for Selective Prediction (JTSP). We find that JTSP outperforms other methods, and also improves the accuracy of each of its modules (CL and DP).

\begin{figure*} 
\centering
\includegraphics[scale=0.265]{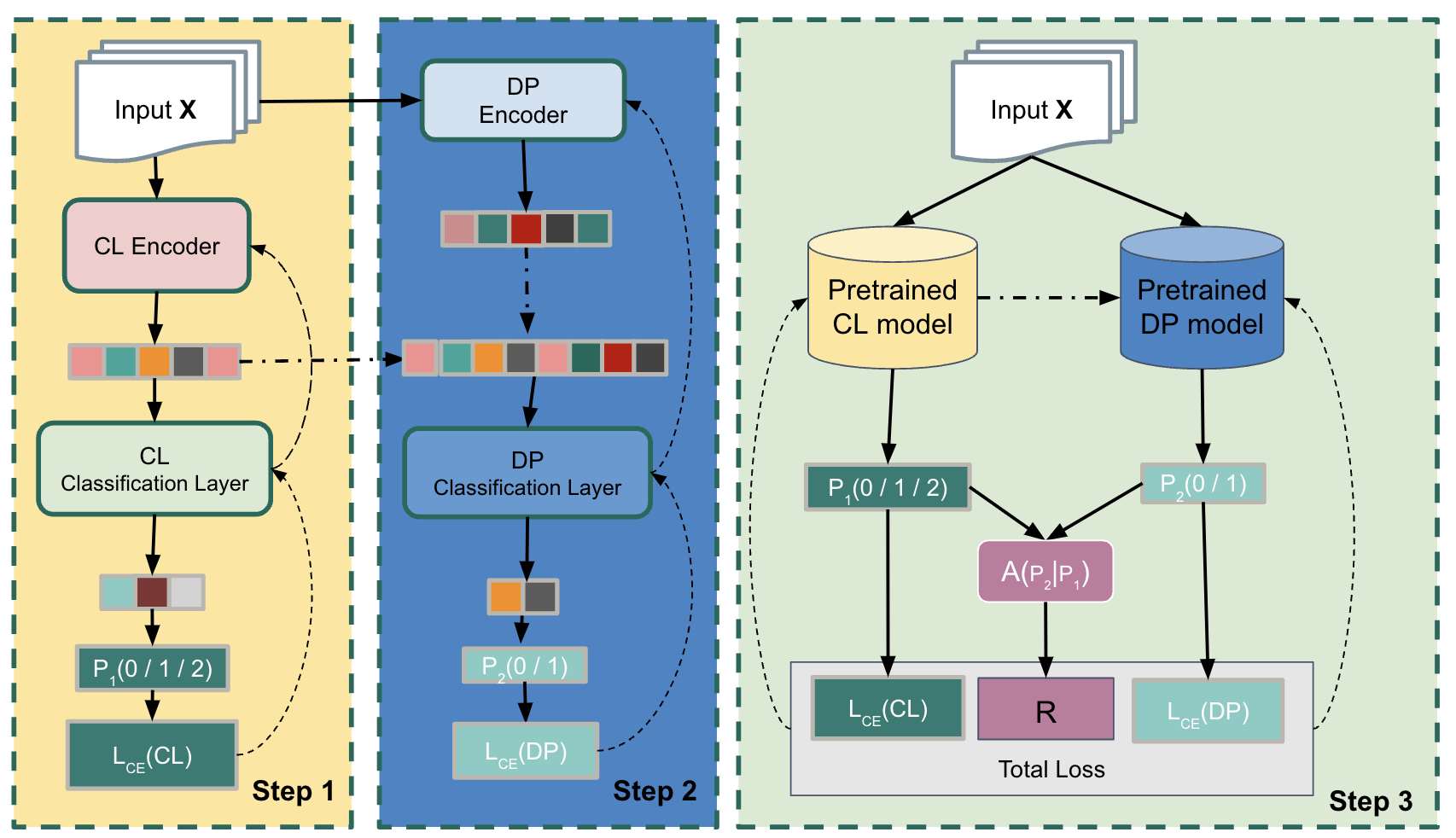}
\caption{The model architecture with three steps. CL is the classification model of the given task, DP is the Deferral Policy model. A is the reward function from equation (\ref{eq:reward}).}
\label{fig:arch}
\end{figure*}

Logically speaking, the ideal SP policy, given a focus on accuracy, would be to defer to a human if and only if an algorithmic decision is incorrect. In practice, this is not only unattainable, there may be situations where a sacrifice in accuracy is worth it, for a greater reduction of human effort. We propose a novel three-step Joint Training for Selective Prediction (JTSP) method, as illustrated in Fig. \ref{fig:arch}. Through joint training, we observe a tradeoff in the two criteria of accuracy and deferral rate, where JTSP produces much higher SP accuracy than baselines, at a modest increase in deferral rate. However, the JTSP training process supports choosing to place a higher weight on one or the other criterion, through the integration of policy gradient methods from reinforcement learning \cite{silver2014deterministic} into the JTSP loss.  During training, the tradeoffs between CL and DP accuracy, and deferral rate, can be controlled through weights on the loss terms, and through a reward signal that manages the co-ordination between the CL and DP.

We identify four main contributions:
\begin{enumerate}[noitemsep,nolistsep]
  \item \textbf{Improved selective prediction}: JTPS improves the SP accuracy, by improving both CL and DP modules.
  \item \textbf{Management of the tradeoff between SP accuracy and deferral rate}: SP systems have two evaluation criteria, accuracy and reduction of human effort (deferral rate). JTSP manages the tradeoff between the two through incorporation of policy gradient reinforcement learning as part of a joint training objective, and a reward signal that co-ordinates the behavior of the CL and the DP.
  \item \textbf{Novel training regimen}: To address the conundrum that the DP cannot be trained until the CL has been trained, the training regimen includes a CL warmup phase, followed by a DP warmup phase, followed by joint training.
  \item \textbf{Test results on two classifiers and four datasets}: We use two types of classifiers for JTSP. Both versions of JTSP outperform strong baselines on four datasets of STEM assessment questions that differ from one another in subject matter, number of questions, and student level.
\end{enumerate}
Here we apply JTSP to assessment of STEM reasoning questions.  The methodology, however, applies to any classification task. The paper includes sections on related work, our approach, experiments and results, discussion, and conclusion.

\section{Related Work}

\textbf{Selective Prediction (SP)}, also known as selective classification, abstention, or the reject option, 
uses model predictions only when given sufficient model confidence \cite{Wiener13, Xin2021}. 
In the simplest case, SP depends on identifying a threshold probability or confidence.

Early work incorporated an explicit reject option among classification labels. \citeauthor{hellman1970nearest} (\citeyear{hellman1970nearest}) pioneered the integration of a reject option in the nearest neighbor rule 
for Bayes' binary classification.
\citeauthor{bartlett2008classification} (\citeyear{bartlett2008classification}) introduced a binary SVM classifier with a reject option designed to pinpoint difficult-to-classify cases, specifically those with conditional probabilities approaching 0.5.
\citeauthor{condessa2013classification} (\citeyear{condessa2013classification}) extended the reject option to an SVM image classifier by leveraging contextual information.
Subsequent studies \cite{yuan2010classification, pillai2013multi, zhang2018reject} expanded the binary reject option framework to encompass multi-class 
scenarios. 
Notably, \citeauthor{pillai2013multi} (\citeyear{pillai2013multi}) is the first work that evaluates the trade-off between accuracy and manual annotation cost/deferral rate. However, it does not 
explore 
machine learning methods to address both.

Recent work on SP has investigated
ways to utilize output probability distributions. 
\citeauthor{hendrycks_baseline_2018} (\citeyear{hendrycks_baseline_2018}) proposed a softmax prediction probability baseline to detect errors and out-of-distribution instances across various tasks in computer vision and NLP. Calibration techniques \cite{garg_will_2021,kamath_selective_2020, Varshney2022settings, jiang_how_2021} harness models' softmax probabilities or hidden representations to 
learn a calibrator model. 
Calibrator
methods typically require a held-out dataset, labeled according to prediction accuracy, to train the calibrator. 
Unlike these methods, 
JTSP eliminates the need for an additional dataset, as it trains the  classification model (CL) and deferral policy (DP) jointly on the same training data. 

\citeauthor{xin2021art} (\citeyear{xin2021art}) emphasized the intrinsic link between SP and confidence estimation, with works like \cite{geifman2017selective} developing selective classifiers that leverage a model's hidden representations to enable confidence level adjustments by users. SP also intersects with out-of-domain (OOD) detection, as explored by \cite{scholkopf1999support, liang2017enhancing}, and aligns with the notion of prediction error detection discussed by \cite{hendrycks_baseline_2018, fort2021exploring}. 
However, a common limitation among the aforementioned approaches, as highlighted by \cite{kamath2020selective}, is that deferral policies based on a model's softmax probabilities often 
suffer from
inherent overconfidence. 
Further, machine learned DPs often overfit the associated CL.
Our work seeks to overcome these limitations by jointly optimizing the representations learned by the CL and DP. 

\textbf{Joint training}, also known as multi-task learning or joint learning, is a 
paradigm where a model is trained simultaneously on multiple related tasks. The key idea 
is that learning to perform multiple tasks at the same time allows a model to share representations 
across
tasks, which can lead to improved generalization 
\cite{caruana1997multitask}. Joint training has 
proved successful across a wide spectrum of machine learning 
problems \cite{collobert2008unified, thung2018brief, zhang2021survey}, typically involving the concurrent training of two models on related tasks to foster improved generalization through shared representations.

A pioneering 
use of joint training 
integrated a deep convolutional network with a Markov random field model, 
achieving significant performance improvements in human body pose recognition over 
state-of-the-art techniques \cite{tompson2014joint}. 
\citeauthor{long2017learning} (\citeyear{long2017learning}) introduced Multilinear Relationship Networks (MRN) to enhance multi-task learning in deep neural networks 
using tensor normal priors to 
improve feature transferrability, and to address issues of negative transfer and under-transfer in different network layers. 
\citeauthor{mao2021joint} (\citeyear{mao2021joint}) demonstrated the efficacy of jointly training two BERT-MRC models with shared parameters for 
machine reading comprehension (MRC). 

Our approach
draws inspiration from joint training strategies mentioned in \cite{zhang2021survey}, which generally involves 
training two models to share the same encoding layer. However, our adaptation involves use of distinct encoders for the CL and DP models, given that the DP decision depends on the class prediction and prediction quality of the CL model. 
The DP output layer uses the concatenation of the representations from the CL and DP encoders, 
which harnesses the benefits of joint training while accommodating the unique demands of SP. 

\section{Approach}

The joint training problem faced here is that the model for the deferral policy has no data to train on until the classifier model can make decisions. Through experimentation, we found that each model needs its own warm up phase, starting with the classifier, and that the loss function for the deferral policy should be rooted in reinforcement learning. The training process is summarized in Algorithm~\ref{algo:train} and Fig.~\ref{fig:arch}.


\begin{algorithm*}[ht] 
\caption{Joint-training for selective prediction from a base classifier (CL) and a deferral policy (DP).
}
\begin{algorithmic}[1]
\State Initialize weights and biases for each layer in the CL and DP modules.
\State \textbf{Warm up:}
\State \hspace{\algorithmicindent} Warm up CL for $n$ epochs with the cross-entropy loss for the classification task $L_{CE}(CL)$.
\State \hspace{\algorithmicindent} Warm up DP for $m$ epochs with the cross-entropy loss for the deferral task $L_{CE}(DP)$.
\State Set learning rate $\eta$ and number of epochs $T$.
\For{$t = 1$ to $T$}
    \For{each training batch $x_i$}
        \State 
        Get the hidden states $h_{i_1}$ (from the CL encoder layer) and $h_{i_2}$ (from the DP encoder layer).
        \State Concatenate the embeddings $h_{i_1}$ and $h_{i_2}$ as a new embedding ${h'}_i$.
        \State Input $h_{i_1}$ into the CL classification layer.    
        \State \hspace{\algorithmicindent} Compute the output CL probability distribution $p_c$ (3D).
        \State Input $h'_i$ into the deferral policy $\pi$.
        \State \hspace{\algorithmicindent} Compute the output DP probability distribution $p_d$ (2D).
        \State Compute the loss for this training batch 
        where R is the policy gradient loss from equation (\ref{eq:reward}):
        \State \hspace{\algorithmicindent} $L = \alpha \cdot L_{ce}(CL) + \beta \cdot L_{ce}(DP) - \gamma R$.
        \State Update the weights and biases using backpropagation.
    \EndFor
    \State Record validation loss and validation accuracy and save checkpoints
\EndFor
\end{algorithmic}
\label{algo:train}
\end{algorithm*}


The first training step 
is a warm-up phase for the classifier.
It follows the 
standard training paradigm for NLP classifiers, 
such as BERT. 
Data examples, e.g., assessment questions and student responses, are first encoded into a hidden representation $h_C$.  A classification layer at the output produces the probability distribution $P_1$, e.g., over the correctness classes. Model parameters $\theta$ are updated through backpropagation of Cross Entropy Loss $L_{ce}(CL)$.

The second training phase temporarily freezes the CL parameters $\theta$ during the deferral policy (DP) warm up. A separate DP encoder receives the same input as the classifier to learn the hidden representation $h_{DP}$. The concatenation [$h_C$,$h_{DP}$] is sent to the DP classification layer to decide whether to defer or not ($P_2$). Again we use cross entropy loss ($L_{ce}(DP)$) to update the DP parameters $\phi$, where the decision to defer is compared with the correctness of the CL decision. The upper bound correctness of DP is to defer if and only if the classifier decision is incorrect.


The third phase commences joint optimization of $\theta$ and $\phi$ by unfreezing $\theta$ and backpropagating from a more complex loss function to update $\theta$ and $\phi$ at every epoch.
In pursuit of achieving superior joint training outcomes and exerting finer control over deferral accuracy, 
we have incorporated policy gradient methods. 
This methodology has a reward function R:


\begin{equation}
\label{eq:reward} 
    \text{R} = \sum \pi(a|s) \times A(s, a) 
\end{equation}
\noindent
where $\pi(a|s)$ represents the likelihood of the action $a$ to defer (1) or not (0), given a particular state $s$ under the deferral policy $\pi$. Here $s$ = $h'_i$, the concatenation of the CL and DP hidden states.

\begin{figure}[b!]
    \centering
    \begin{tabular}{c c c}
            & \multicolumn{2}{c}{DP} \\ 
     CL     &  \multicolumn{1}{|c|}{-Defer} &  \multicolumn{1}{|c|}{+Defer} \\\hline
     Cor.   &  \multicolumn{1}{|c|}{$a$}  &  \multicolumn{1}{|c|}{$b$}  \\\hline
     Incor. &  \multicolumn{1}{|c|}{$c$}  &  \multicolumn{1}{|c|}{$d$}   \\\hline
    \end{tabular} 
    \\
    \vspace{.2cm}
    \begin{tabular}{l}
    $A = [a, b, c, d]$ where \\
    $\forall x \in [a, b, c, d], x \geq 0,$  sum$([a, b, c, d]) = 1$
    \end{tabular}
    \caption{Four-valued reward signal $A$. The desirable outcomes $a$ and $d$ should have higher values than the undesirable outcomes $c$ and $d$.}
    \label{fig:reward-signal}
\end{figure}

To 
jointly optimize 
classification accuracy and deferral accuracy, we 
designed a reward signal $A(s, a)$ for DP decisions (actions) that has distinct scalar values for each of the four outcomes of correctness of the classifier by correctness of the decision to defer, as shown in Fig.~\ref{fig:reward-signal}. To facilitate interpretation of the performance of the reward signal $A$, we constrain the four values to be positive, and to sum to 1. In general, designing an optimal reward signal for reinforcement learning is challenging. As discussed in the next section, we found best performance with $a > d >> (b | c)$.

The final loss for the joint training phase is a weighted 
sum of the cross-entropy losses from both models and the policy gradient loss (equivalent to negative R).
\begin{equation}
\label{eq:loss} 
    L_{JTSP} = \alpha \cdot L_{ce}(\text{CL}) + \beta \cdot L_{ce}(\text{DP}) - \gamma R 
\end{equation}


At inference time, selective prediction for each input $x$ involves two decisions, one made by the CL based on its learned representation of $x$ ($h_{CL}$), and the DP's decision based on concatenating its learned representation of $x$ ($h_{DP}$) with $h_{CL}$.

\section{Experiments and Results}
\label{sec:exp}

\begin{table*}[ht]
\centering
\caption{
Comparison of selective prediction baselines with JTSP, on four short answer datasets. DP is the accuracy/F1 of the deferral policy; SP  is the overall accuracy/F1; DR is the deferral rate (lower is better). For each model and metric, the best value within a column is in boldface. For each dataset, an asterisk is on the best SP.
}
\label{tab:comprehensive_comparison}

\begin{subtable}{\textwidth}
\centering
\small
\caption{Performance on BEETLE and SciEnts}
\label{tab:beetle_sci}
\begin{tabular}{l | l | ccr | ccr }
\hline
CL & DP  & \multicolumn{3}{c|}{BEETLE} & \multicolumn{3}{c}{SciEnts}   \\\cline{3-8}
 &  & DP & SP & DR  & DP & SP & DR\\
\hline
SFRN & None & NA & 79.35 / 70.73 & NA  & NA &  75.11 / 65.35   & NA \\
& Thresh. & 79.50 / 50.20 & 80.74 / 72.68 & \textbf{2.63} & 74.96 / 45.57 &  75.85 / 65.14& \textbf{1.63} \\
& LR &  80.00 / 59.92 & 83.89 / 77.67 & 8.64  & 75.07 / 47.76 & 75.74 / 66.52 & 3.23 \\
& Policy  & 79.66 / 58.26 & 83.39 / 76.98 & 7.76 & 75.11 / 47.08  &  76.78 / 68.94 & 3.37 \\
& JTSP$_{CE}$   & 79.89 / 56.87 & 83.58 / 77.64 & 8.04 & 76.61 / 55.85 & 79.41 / 73.60 & 7.48 \\
& JTSP & \textbf{81.02 / 61.47} & \textbf{*85.33 / 79.63} & 8.20 & \textbf{76.67  / 58.44} &  \textbf{*80.00 / 73.78} & 8.56 \\
 \hline
BERT & None & NA  & 77.8 / 70.35 & NA  & NA & 73.62 / 60.75  & NA \\
& Thresh. & 77.17 / 52.31 & 80.12 / 74.24 & 5.43 & 73.77 / 45.06 & 74.37 / 62.55 & \textbf{1.33} \\  
& LR  & 76.32 / 50.45 & 78.47 / 70.65 & \textbf{4.77} & 73.76 / 45.38 & 74.67 / 61.97 & 1.63 \\
& Policy  & 77.37 / 58.64 & 82.92 / 77.99 & 10.68 &  \textbf{75.43} / 48.11 &  76.53/ 67.20  & 3.31 \\
& JTSP$_{CE}$   & 79.35 / 62.73 & 83.46 / 77.37 & 9.27 &  75.04 / 45.07 & 75.96 / 65.14  &  4.78\\
& JTSP & \textbf{80.90 / 64.03} & \textbf{85.29 / 80.08} & 10.71 & 74.70 / \textbf{52.70} &  \textbf{77.37 / 69.56} &  5.93\\
\hline
\end{tabular}
\end{subtable}

\vspace{1em} 

\begin{subtable}{\textwidth}
\centering
\small
\caption{Performance on Mid-PHYS and ISTUDIO}
\label{tab:midphys_istudio}
\begin{tabular}{l | l | ccr | ccr }
\hline
CL & DP & \multicolumn{3}{c|}{Mid-PHYS} & \multicolumn{3}{c}{ISTUDIO} \\\cline{3-8}
 &  & DP & SP & DR & DP & SP & DR \\
\hline
SFRN & None & NA &  79.5 / 78.38  & NA & NA &  83.19 / 80.37   & NA \\
& Thresh. & 79.65 / 49.54 &  80.67 / 80.01 & 2.65 & 16.07 / 13.85 & 100.0 / 100.0 & 100.0 \\  
& LR  & 78.88 / 49.26 & 80.37 / 79.94 & 2.74 & 73.22 / 35.68 & 85.32 / 83.36 & 5.52 \\
& Policy  & 79.83 / 49.08 & 80.75 / 79.69 & \textbf{3.50}  & 83.25 / 50.73 & 85.03 / 82.95  & 2.49 \\
& JTSP$_{CE}$  & \textbf{80.97 / 56.84} & 82.49 / 81.63 & 4.61 & 84.95 / 45.93 &  84.95 /83.40 & \textbf{0.03} \\
& JTSP  & 80.53 / 56.49 &  \textbf{*83.33 / 82.49} & 5.90 & \textbf{85.07 / 52.18} & \textbf{*85.83 / 84.29} & 1.84 \\
 \hline
BERT & None  & NA  & 78.76 / 78.04 & NA  & NA  &  81.28 / 79.88  & NA  \\
& Thresh. & 80.23 / \textbf{57.87} & 82.44 / 81.87 & 5.90 & 82.38 / 45.17 & 82.69 / 81.13 & 0.30 \\  
& LR  & 80.75 / 57.28 & 81.69 / 81.27 & 5.64 & 80.91 / 40.45 & 81.45 / 80.39 & 0.38 \\
& Policy  & 79.09 / 47.67 & 79.57 / 78.91 & 5.45 &  82.70 / 45.89  & 82.85 / 81.29 & \textbf{0.23} \\
& JTSP$_{CE}$  & \textbf{80.64} / 53.93 & 82.41 / \textbf{82.16} & \textbf{5.02} & 82.39 / 45.94 &  82.77 / 81.45 & 0.69 \\
& JTSP & 79.68 / 56.65 & \textbf{82.63} / 81.73  & 5.42 & \textbf{82.81 / 46.17} &  \textbf{83.35 / 81.98} & 0.77 \\
\hline
\end{tabular}
\end{subtable}

\end{table*}

The experiments compare JTSP incorporating two classifiers that have different architectures. BERT~\cite{devlinEtAl19} is a general-purpose, high-performing, pre-trained encoder language model based on transformers. Here we use HuggingFace's \href{https://huggingface.co/docs/transformers/v4.37.2/en/model_doc/bert#overview}{BERT AutoModelForSequenceClassification} with fine-tuning. The second classifier, SFRN, was developed specifically for short answer assessment, consisting of a relation network initialized with BERT encodings. SFRN had SOTA performance on two of the datasets used here~\cite{Li2021}. We compare the two classifiers when used as the encoders for CL under five selective prediction settings, using four datasets of students' written answers to STEM questions. In all conditions, RoBERTa \cite{liu2020roberta} is used as the DP encoder, which performed better than BERT. On all four datasets, SFRN as part of a full JTSP system performs best. Further, both classifiers improve in the joint training context. The remainder of the section presents the three SP conditions, the four datasets, results on selective prediction, and results of the classifiers with and without joint training.

The five selective prediction settings are labeled as: Thresh., LR, Policy, JTSP$_{CE}$ and JTSP. For the Thresh. condition, we select a probability threshold for the maximum probability class, based on the validation sets as in \cite{hendrycks_baseline_2018}: that is, the threshold should produce SP accuracy greater than the CL accuracy alone. The other approach we know that learns a deferral policy is from~\cite{li2023learning}, also in the domain of assessment of short answer questions. It presents a logistic regression policy that uses these features from the short answer classifier: the predicted class, the softmax probabilities for each class, the classifier's training accuracy on the question. Our LR condition is a logistic regression that uses the same features. In the Policy condition, the classifier for student answers and deferral policy classifier are trained separately, each with their own cross entropy loss. JTSP$_{CE}$ represents joint training but without the use of the policy gradient loss term shown in line 15 of Algorithm \ref{algo:train}). The JTSP condition stands for the full joint training algorithm depicted in Fig.~\ref{fig:arch} and Algorithm~\ref{algo:train}.

Four datasets of STEM questions were used here. Two are from the SemEval-2013 Task 7 dataset \cite{dzikovska2013semeval}.\footnote{The original download link is no longer available, but the authors can share this dataset on request.} BEETLE consists of about 6,000 undergraduate students' responses to questions about electricity and electronics. SciEntsBank (henceforth SciEnts) \cite{nielsen2008annotating} consists of about 10,000 answers to assessment questions across 15 science domains from students in grades 3-6, with inter-annotator agreement of 72.8 Kappa. For BEETLE and SciEnts, we use the 3-way correctness classes, unseen answers, where the classes are \textit{Correct, Incorrect} and \textit{Contradictory}. Mid-PHYS, derived from a decade's worth of learning design studies in Wisconsin middle schools, features pre-/post-test data aimed at assessing middle school students' understanding of physics concepts, with the questions being expert-validated to accurately measure comprehension of physics relationships. It has 11,245 constructed responses to 55 physics questions. ISTUDIO\footnote{Available at doi:10.26208/JFMP-V777.} consists of responses from 1,935 undergraduate students to six 2-part statistics questions totaling over 6,000 responses, with a Fleiss Kappa score for three annotators of 0.70. For Mid-PHYS and ISTUDIO, there are three correctness classes: \textit{Correct, Partially Correct} and \textit{Incorrect}.

\begin{table*}[ht]
\centering
\caption{Comparison of two classifiers trained as standalone systems or as part of JTSP$_{CE}$ or JTSP, on four short answer datasets. Cell values are accuracy and F1 (Acc/F1).
}
\begin{tabular}{l | l | cccc }
\hline
CL & DP & BEETLE & SciEnts & ISTUDIO & Mid-PHYS  \\
\hline
SFRN & None & 79.35 / 70.73 & 75.11 / 65.35 & 83.19 / 80.37 &  79.5 / 78.38 \\
& JTSP$_{CE}$ & 80.16 / 72.26 & {\bfseries 75.22 / 66.06} & {\bfseries 84.95 / 83.34} & 79.39 / 75.25 \\
& JTSP & {\bfseries 80.43 / 72.42} & 74.78 / 63.51 & 84.76 / 83.06 & {\bfseries 80.24 / 79.24} \\
\hline
BERT & None & 77.8 / 70.35 & 74.52 / 64.09  & 81.28 / 79.88 & 78.76 / 78.04   \\
& JTSP$_{CE}$ & 77.53 / 68.98 & {\bfseries 75.37 / 64.23} &  82.62 / 81.31 & {\bfseries 81.16 / 80.25} \\
& JTSP & {\bfseries 79.19 / 71.04} & 74.11 / 61.34 & {\bfseries 83.19 / 81.84} & 79.24 / 78.24 \\
\hline
\end{tabular}
\label{tab:classifiers}
\end{table*}

During the training phase, we first warm up each model for 10 epochs. Then we tune the JTSP loss hyper-parameters \([\alpha, \beta, \gamma]\) (see equation \ref{eq:loss}), selecting their values based on experience gained during joint training. 
The initial setting for each dataset began with \([1, 1, 1]\), followed by a systematic exploration aimed at enhancing performance. 
We observed that optimal settings for \(\alpha\), \(\beta\), and \(\gamma\) differed across datasets. 
The experiments reported here use the following values for  \([\alpha, \beta, \gamma]\): BEETLE \([0.01, 0.01, 15]\); SciEnts \([0.1, 0.1, 10]\); Mid-PHYS \([0.1, 0.1, 1]\); ISTUDIO \([1, 1, 1]\).

Table \ref{tab:comprehensive_comparison} presents the JTSP results, broken down by dataset and by short answer classifier (SFRN, BERT).  The three results columns are accuracy/F1 scores for the deferral policy classifier (DP), accuracy/F1 scores for the full selective prediction system (SP), and the deferral rate (DR).  The results for the SFRN and BERT classifiers are in Table~\ref{tab:classifiers}, and discussed below. In all cases, SP results are on test sets where, if the deferral policy decision is to defer, the ground truth labels are used.

We now discuss the use of a learned deferral policy using a neural network, with and without joint training, compared with Thresh. and LR. For each dataset and CL model (SFRN or BERT), the best accuracy, F1 and deferral rate are in bold face. For each dataset, the best SP result has an asterisk. The threshold method always has the lowest SP accuracy, given a particular CL model, apart from ISTUDIO where the threshold picked on the validation set led to 100\% human effort, therefore perfect accuracy. This highlights the risk of selecting a deferral threshold based on softmax. JTSP/SFRN has the highest SP accuracy/F1 on all datasets, generally with the highest DP accuracy/F1 as well (apart from Mid-PHYS). At the same time, it often has a higher deferral rate than other SFRN approaches, but deferral rates are very low overall, surpassing 10\% only for Policy/BERT. JTSP/BERT generally outperforms other BERT-based SP methods on SP accuracy/F1, and comes close to JTSP/SFRN on BEETLE and Mid-PHYS. Joint training (both JTSP conditions) always surpasses Policy, where the CL and DP modules are trained independently. LR and Policy are often close, but LR/BERT outperforms Policy on BEETLE and Mid-Phys. In sum, results show that JTSP leads to superior SP accuracy compared with other deferral methods, for both CL models. JTSP/SFRN outperforms JTSP/BERT, possibly because SFRN was developed specifically for short answer assessment.


Considering the DP module in greater detail, DP accuracy tends to increase in each next condition, for both CL variants, apart from Mid-PHYS, where the JTSP$_{CE}$ variant has better DP accuracy for both CL models. Often, the Thresh. method has the lowest deferral rate, but at a sacrifice in SP accuracy. However, there is no consistent trend for the deferral rates across models and datasets.  This might simply reflect deferral rates that are so low, that small changes in the DP might have greater impact on some datasets, e.g., SciEnts, compared with others, e.g., Mid-PHYS. As we discuss in the next section, we speculate that it is more difficult to optimize the hyper-parameters and reward signal for the DP classifier, because its performance is bounded by the performance of the CL classifier.

Table \ref{tab:classifiers} shows that with joint training, the CL classifiers improve, thus both tables show that joint training benefits both the CL and DP modules. For BEETLE and Mid-PHYS, SFRN improves most under the JTSP condition, and for the other two datasets, it improves most with JTSP$_{CE}$. BERT improves most with JTSP$_{CE}$ on SciEnts and Mid-PHYS, and improves most with JTSP on BEETLE and ISTUDIO.  



Our results suggest that 
the synergy between the CL and DP modules 
merits further investigation, including choice of encoders. Moreover, the observed improvements highlight the potential for developing more advanced DP models and training strategies that could further enhance the performance of SP systems and the CL models they use across a variety of tasks and datasets.

\section{Discussion}
\label{sec:discussion}



\begin{figure*}[t!]
    \centering
    \begin{subfigure}[b]{0.5\textwidth}
        \centering
        \includegraphics[scale=0.33]{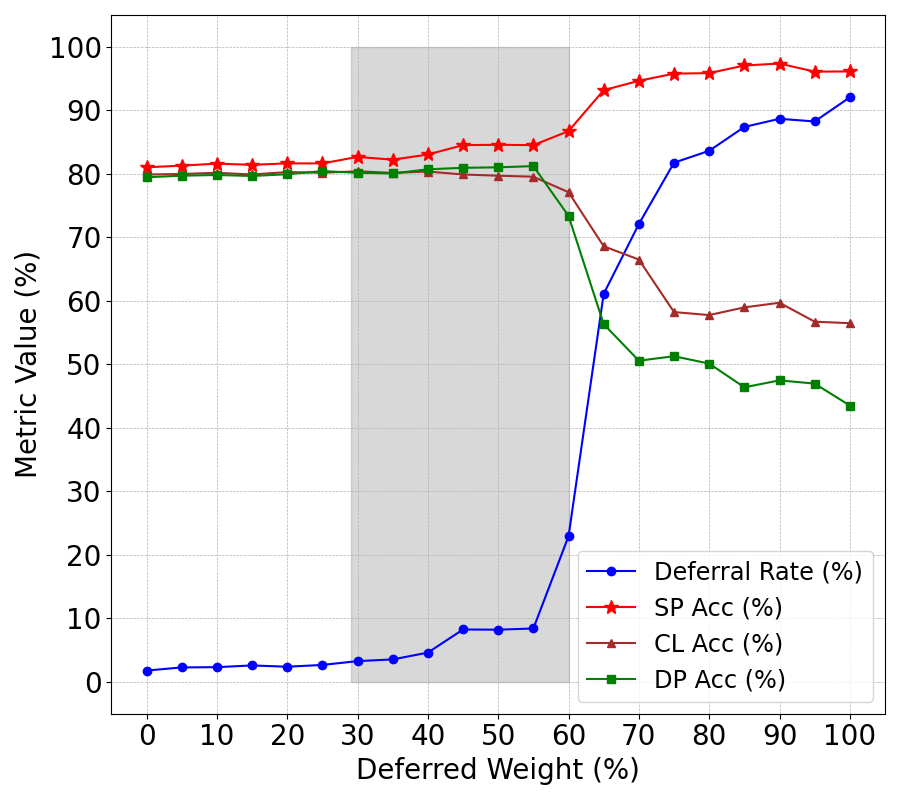}
        \caption{JTSP with SFRN classifier.}
        \label{fig:beetle_SFRN}
    \end{subfigure}%
    ~ 
    \begin{subfigure}[b]{0.5\textwidth}
        \centering
        \includegraphics[scale=0.33]{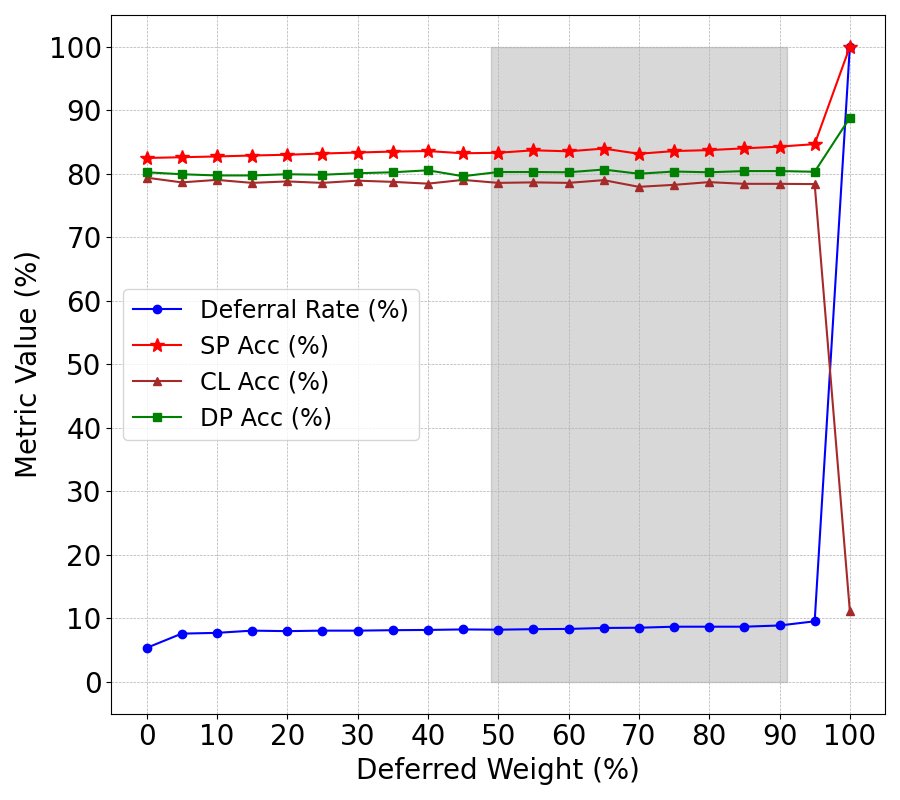}
        \caption{JTSP with BERT classifier.}
        \label{fig:beetle_BERT}
    \end{subfigure}
    \caption{
    Sensitivity of four metrics to deferral weight ($d$ in the reward signal $A$) on the BEETLE dataset: deferral rate (blue), SP accuracy (red), CL accuracy (brown) and DP accuracy (green). 
    }
    \label{fig:beetle}
\end{figure*}

On the four datasets tested here, JTSP improves SP accuracy over the baseline SP methods. JTSP achieves this result from a joint training method where the CL and DP learn distinct representations of the input, as noted in the preceding sections, while training the DP classifier on the concatenation of both representations. The improved SP accuracy derives from improvements to both the CL and DP learning over the Policy condition, where they are trained independently.
In addition,
the increased deferral rates associated with higher 
SP accuracy represent a strategic trade-off. 
This tradeoff is partly managed by the reward signal $A$.

The reward signal takes into account the two DP actions, to defer or not, and the two conditions for each, the action is correct or incorrect.  The results presented above used $A = [0.5, 0.1, 0, 0.4]$, which performed well across all conditions, in comparison to hundreds of other values. It gives the highest weights to $a$ and $d$, the cells representing the ideal cases of a correct CL decision and no deferral ($a$), and deferring given an incorrect CL decision ($d$).  It also weights CL correctness ($a+b$) slightly higher than the action to defer ($b+d$). Higher values of $d$ typically raise the deferral rate while lowering CL accuracy, and higher values of $a$ tend to lower the deferral rate while increasing CL accuracy. In this way, $A$ manages the tradeoff between overall accuracy and the deferral rate. However, the change in performance as $A$ changes is complex.

If it were possible to continue improving the accuracies of the CL and DP modules in DP, then presumaby the SP accuracy could continue to grow while keeping the deferral rate low. This is not the behavior we observe, however, when we gradually change $A$. To avoid too many confounding factors, we conducted a sensitivity analysis on $A$ where $b$ and $c$ are kept at zero, and we vary $d$; in this setting, $a$ is constrained to be $1-d$.  For convience, we refer to $d$ as the deferral weight. We find large differences in sensitivity to $d$ across four metrics, depending on the dataset and model.

Fig.~\ref{fig:beetle} shows two sensitivity plots with $d$ on the $x$-axes, and accuracy or rate on the $y$-axes. We varied $d$ from 0 to 1 using a step size of 0.05, and plotted change in deferral rate (DR; blue), SP accuracy (red), CL accuracy (brown), and DP accuracy (green). An ideal SP system would have high SP accuracy (red) and low DR (blue). We show JTSP on BEETLE using SFRN (Fig.~\ref{fig:beetle_SFRN}) or BERT (Fig.~\ref{fig:beetle_BERT}), illustrating that DR rises more slowly for the SFRN variant, while for the BERT variant it remains very low until a sudden sharp increase at $d=0.95$.  Within the greyed region of Fig.~\ref{fig:beetle_SFRN}), we see that all metrics begin to increase at $d=0.30$. At $d=0.40$, DR starts to increase, yet the accuracies of SP, CL and DP are still slowly increasing. In the range $0.45 \leq d \leq 0.55$, DP accuracy increases as $d$ increases, while SP accuracy remains unchanged due to diminished CL accuracy. This highlights the difficulty of joint optimizion of CL and DP. At $d=0.55$, there is a sharp increase in DR and it continues to rise steeply, concurrent with a sharp drop in CL and DP accuracies.

For the BERT variant, there is far less sensitivity to $d$: all accuracies remain relatively constant for values of $d \geq 0.50$, and all metrics change dramatically at $d=0.95$. Oddly, at the point where DR suddenly jumps from around 10\% to nearly 90\%, the DP accuracy improves from around 80\% to nearly 90\%. In sum, as $d$ changes, JTSP joint training clearly has less influence on altering the BERT representations, which then leads to less influence on the RoBERTa representations used in the DP. In this setting, BERT's stability is somewhat of a disadvantage, given that the BERT variant of JTSP performs less well than the SFRN variant on all datasets. As mentioned in the preceding section, the disparity between the behavior of JTSP with SFRN versus BERT may be due to the fact that SFRN was designed specifically for student's responses to assessment questions, whereas BERT was developed as a general-purpose model. Two questions this raises for future work are how to select models that are more susceptible to improved representation learning, and how to avoid the sudden performance \textit{``cliff''} where CL or DP accuracy suddenly degrades and DR suddenly increases.

\section{Conclusion}

We have presented JTSP, a method to jointly learn a classifier (CL) and a deferral policy (DP) for selective prediction (SP) that relies on a novel architecture, training procedure and loss function. Instead of learning a single representation to meet multiple objectives, JTSP learns different representations to support each of two decisions. On four student assessment datasets with diverse characteristics, JTSP outperforms baselines based on prior work.  Previous SP methods aimed to improve confidence estimation of the classifier decision, given a fully trained classifier.  We show that in the context of JTSP, both the CL and the DP classifiers improve their representation learning.  

The overall behavior of JTSP is highly dependent on the reward signal.  A sensitivity analysis of deferral weight $d$, based on a somewhat more constrained reward signal, demonstrated that sensitivity to $d$ varies dramatically, depending on the choice of model for the CL classifier. The SFRN variant of JTSP is more susceptible to improved representation learning. A potentially fruitful direction for future work would be learn a reward signal that better customizes JTSP to a new dataset, and where the tradeoff between selective prediction accuracy and deferral rate can be better controlled.


\section{Limitations}

The main limitation of JTSP is that the reward signal that has three degrees of freedom, thus is difficult to optimize. 
We are optimistic that further investigation could address this limitation through a novel method to learn the reward signal. We found that across datasets, the optimal hyperparameters and performance of JTSP with the two loss functions varied in unexpected ways. This variation also argues for the potential benefit of an approach that would learn the reward signal. 

A similar limitation is that we allowed the loss hyperparameters to vary independently of one another. A very large number of experiments was needed at the outset of the project to develop the appropriate joint training method. We found that fewer constraints on the values of $\alpha, \beta$ and $\gamma$ (meaning a much larger search space), led more rapidly to combinations that improved performance significantly. However, we believe it could be useful to constrain these hyperparameters in future work.

Selective prediction could be used for any classification task, and with any classifier architecture. Here we tested only one task, correctness classification of students' answers to open-ended STEM questions; only two CL architectures, BERT and SFRN; only one encoder for the DP module. Early experiments showed that using RoBERTa for DP and BERT or SFRN for CL outperformed other combinations, such as BERT in both modules. However, many combinations are possible. It is an open-question how well JTSP would perform in other settings.


\bibliographystyle{acl_natbib}

\end{document}